\documentclass{article}

\usepackage{microtype}
\usepackage{graphicx}
\usepackage{subfigure}
\usepackage{hyperref}
\usepackage{booktabs} 

\usepackage{hyperref}

\usepackage[accepted]{icml2025}

\usepackage{amsmath}
\usepackage{amssymb}
\usepackage{mathtools}
\usepackage{amsthm}
\usepackage{multirow}
\usepackage[normalem]{ulem}
\useunder{\uline}{\ul}{}
\usepackage[capitalize,noabbrev]{cleveref}

\theoremstyle{plain}

\theoremstyle{definition}

\theoremstyle{remark}

\usepackage[textsize=tiny]{todonotes}

\icmltitlerunning{Towards Scalable and Structured Spatiotemporal Forecasting}

\begin{document}

\twocolumn[
\icmltitle{Towards Scalable and Structured Spatiotemporal Forecasting}



\icmlsetsymbol{equal}{*}

\begin{icmlauthorlist}
\icmlauthor{Hongyi Chen}{yyy}
\icmlauthor{Xiucheng Li}{yyy}
\icmlauthor{Xinyang Chen}{yyy}
\icmlauthor{Jing Li}{yyy}
\icmlauthor{Kehai Chen}{yyy}
\icmlauthor{Liqiang Nie}{yyy}
\end{icmlauthorlist}

\icmlaffiliation{yyy}{Harbin Institute of Technology, Shenzhen, China}

\icmlcorrespondingauthor{Xiucheng Li}{lixiucheng@hit.edu.cn}

\icmlkeywords{Machine Learning, ICML}

\vskip 0.3in
]



\printAffiliationsAndNotice{} 

\begin{abstract}
In this paper, we propose a novel Spatial Balance Attention block for spatiotemporal forecasting. 
To strike a balance between obeying spatial proximity and capturing global correlation, we partition the spatial graph into a set of subgraphs and instantiate Intra-subgraph Attention to learn local spatial correlation within each subgraph; to capture the global spatial correlation, we further aggregate the nodes to produce subgraph representations and achieve message passing among the subgraphs via Inter-subgraph Attention. Building on the proposed Spatial Balance Attention block, we develop a multiscale spatiotemporal forecasting model by progressively increasing the subgraph scales. The resulting model is both scalable and able to produce structured spatial correlation, and meanwhile, it is easy to implement. We evaluate its efficacy and efficiency against the existing models on real-world spatiotemporal datasets from medium to large sizes. The experimental results show that it can achieve performance improvements up to 7.7\% over the baseline methods at low running costs.
\end{abstract}

\section{Introduction}
\label{sec:intro}
Multivariate time series data records physical quantities of interest of $N$ variates spanning over $T$ time steps, and spatiotemporal data is a particular type of multivariate time series data in which each variate is equipped with a spatial coordinate in a metric space.  Spatiotemporal forecasting finds widespread use in traffic management, meteorology studies, and (wind) solar energy prediction. Inspired by the milestone achievements of Transformer in natural language processing and computer vision~\cite{DBLP:conf/nips/BrownMRSKDNSSAA20, DBLP:conf/iclr/DosovitskiyB0WZ21}, the recent advancements of multivariate time series are mostly dedicated to designing time series analogs to model long-range time dependencies~\cite{zhou2022fedformer, patchtst2023}. Since the underlying dynamics of spatiotemporal data are jointly characterized by the spatial correlation and temporal dependencies, and thus it is equally important to effectively capture the correlation from the spatial dimension, as evidenced by the success of iTransformer~\cite{liu2024itransformer}.

The spatial correlation can be interpreted as the topological relationship among different variates. Aside from the general topology shared by multivariate time series data, spatiotemporal data also possesses the spatial proximity induced by the spatial coordinates, which serves as a strong prior and has a prominent impact on the topology generation. This is summarized by Tobler's first law of geography---everything is related to everything else, but near things are more related than distant things~\cite{miller2004tobler}. The spatial proximity can be described by a graph (referring to as spatial graph), in which each variate is a node, and an edge is drawn between two nodes if their spatial distance is less than a predefined threshold. 

In the spirit of Tobler's first law, DCRNN~\cite{li2017diffusion} encodes the prior into the model by running graph convolution on the spatial graph to make the forecasting, which obtains pronounced performance gains. Despite the achievement, it makes the distant node pairs very hard to exchange information due to the limitations (such as over-smoothing and over-squashing) of message-passing paradigm~\cite{DBLP:conf/icml/Ma0LRDCTL23}. However, Tobler's second law of geography points out that the phenomenon external to a geographic area of interest affects what goes on inside~\cite{miller2004tobler}, and many works~\cite{STWA, DBLP:journals/tkde/GuoLWLC22} have also confirmed that the distant nodes could also manifest strong correlations. To sidestep this, the proposals~\cite{bai2020adaptive, wu2020connecting, shang2021discrete, jiang2024sagdfn, liu2024itransformer} simply drop the spatial graph and learn the spatial correlation with attention mechanism in an end-to-end manner. As the attention mechanism enables every two-node pair to communicate instantaneously, it can effectively resolve the distant message-passing issue.  However, such methods introduce two drawbacks: 1) the cost of generating pairwise spatial correlation is $\mathcal O(N^2)$ in terms of both computational and memory complexity; 2) the learned spatial correlation matrix is unstructured (lacks spatial sparsity) and contains a large fraction of trivial nonzero entries (noise), the accumulated noise is likely to impair the forecasting performance when $N$ is large. 

To address the two limitations, we propose a novel Spatial Balance Attention block that respects the first law of geography while also permitting long-distance message passing in this paper. The high-level idea is to partition the spatial graph into a set of subgraphs and instantiate self-attention to learn local spatial correlation within each subgraph---Intra-subgraph Attention; to capture the global spatial correlation, we further aggregate the nodes to produce subgraph representations and exchange information among the subgraphs via self-attention again---Inter-subgraph Attention. Since the entire block is differentiable, the message passed from distant nodes can be backpropagated through Inter-subgraph Attention to their correlated nodes in a parsimonious manner. 
Moreover, we stack the proposed Spatial Balance Attention blocks with residual and gradually increase the subgraph scales to develop our eventual spatiotemporal forecasting model. Such a design brings the following two appealing features: 1) our proposed method adopts the spatial graph to encourage local spatial information exchange, and thus it can yield sparse structure and reduce both the computational and memory burdens; 2) it permits message passing between distant node pairs in a parsimonious way, and thus it is capable of capturing global spatial correlation without incurring extra noise. To summarize:
\begin{itemize}
    \item We propose a novel Spatial Balanced Attention block for spatiotemporal forecasting that strikes a balance between obeying spatial proximity and capturing global correlation. 
    \item Building on the proposed block, we develop a multiscale spatiotemporal forecasting model by gradually increasing the subgraph scales. The resulting model is scalable and can produce structured spatial correlation.
    \item Our proposed method is effective yet easy to implement. We evaluate its efficacy and efficiency on real-world spatiotemporal datasets from medium to large sizes. It can achieve performance improvements up to 7.7\% over baseline methods while maintaining low running costs. 
\end{itemize}

\section{Related work}
\label{sec:related-work}
\subsection{Temporal Dependency Modeling}
Spatial-temporal series forecasting belongs to multivariate time series analysis and has been studied for decades. Early deep sequential models commonly employed in this field include recurrent neural networks (RNNs) \cite{zhao2017lstm,lai2018modeling,salinas2020deepar} and convolutional neural networks (CNNs) \cite{bai2018empirical,hewage2020temporal,wu2022timesnet}. Benefiting from the wide receptive field of attention mechanism, Transformer \cite{vaswani2017attention} based models have been widely acknowledged in time series forecasting to capture the long-term temporal dependencies adaptively \cite{wu2021autoformer,zhou2021informer,zhou2022fedformer,patchtst2023}. Furthermore, multi-layer perceptrons (MLPs) have also been introduced \cite{zeng2023transformers,challu2023nhits,wang2024timemixer} for time series forecasting, demonstrating that even a simple model can effectively extract strong temporal periodic dependencies.

\subsection{Spatial Correlation Modeling}
Different from multivariate time series forecasting, spatial-temporal series forecasting pays more attention to spatial correlation except for the temporal dependencies. The advancement of graph neural networks (GNNs) offers an effective way to model unstructured spatial adjacency correlation. In the spirit of Tobler’s first law of geography, DCRNN \cite{li2017diffusion}, STGCN \cite{yu2017spatio} and TGCN \cite{zhao2019t} leverage a spatial graph based on real-world distance and propose to fuse the local spatial information via graph convolution operation. However, it makes the distant node pairs very hard to exchange information due to the limitations of message-passing paradigm~\cite{DBLP:conf/icml/Ma0LRDCTL23}, violating Tobler’s second law of geography. Subsequently, adaptive GNNs-based methods have been proposed to solve this problem. GWNET \cite{wu2019graph}, AGCRN \cite{bai2020adaptive}, MTGNN \cite{wu2020connecting} and CrossGNN \cite{huang2023crossgnn} learn a representation for each series and then generate the correlation graph through computing the pairwise interaction of the node representations. GTS \cite{shang2021discrete} and STEP \cite{shao2022pre} directly learn a discrete graph based on history time series. Benefiting from naturally constructing a fully connected graph with learnable edge weights, the self-attention mechanism is also a commonly adopted method for capturing global and dynamic spatial correlation \cite{zheng2020gman,jiang2023pdformer,liu2024itransformer,wang2024card}. 
Nevertheless, the learned spatial correlation matrix in such methods is unstructured (lacks spatial sparsity) and contains a large fraction of trivial nonzero entries (noise). The accumulated noise is likely to impair the forecasting performance when \(N\) is large. Moreover, they require \(\mathcal{O}(N^2)\) computational complexity, impeding their application in large-scale datasets. In addition, some researchers attempt to capture dynamic spatial correlation by learning evolving spatial graph, albeit with high computational and memory costs \cite{shao2022decoupled,chenstructured}.

\subsection{Scalable Spatial Correlation Modeling}
Several researchers have focused on developing scalable spatiotemporal forecasting methods to accommodate larger datasets. Model-agnostic methods introduce a graph partition approach that decomposes large spatial graph into smaller ones and conduct experiments through independent training \cite{mallick2020graph} or continual learning \cite{wangmake}. In contrast, existing scalable models can be roughly categorized into the following four paradigms: computing spatial correlation in advance, performing linearized spatial correlation computation, sampling nodes of the spatial graph and mixing channels in spatial dimensions. SGP \cite{cini2023scalable} and SimST \cite{liu2024reinventing} precompute graph convolutions for node features and decouple the spatial correlation modeling from the training process. Nevertheless, these methods might exhibit reduced effectiveness owing to fixed representation in the input space. BigST \cite{han2024bigst} and Sumba \cite{chenstructured} develop a linearized spatial convolution operator to reduce complexity but fail to produce structured spatial correlation due to the low rank approximation. HGMTS \cite{kim2023hierarchical} reduces the computational overhead of self-attention by identifying pivotal query nodes and sampling associated key nodes. SAGDFN \cite{jiang2024sagdfn} integrates a significant neighbors sampling module and only captures spatial correlation with significant neighbors. However, these sampling methods struggle to preserve the node's local structural information. In addition, the channel mixing approach is a widely used method to improve scalability \cite{zhang2023crossformer,yeh2024rpmixer,han2024softs}. It aggregates messages from all dimensions and distributes the received messages among dimensions through a routing mechanism. Even though it avoids the quadratic asymptotic complexity, it introduces a large constant term and exhibits poor scalability in practice.

\section{Preliminaries}

\textbf{Spatiotemporal Series Data}.
Spatial-temporal series data refers to multivariate time series data in which each series is equipped with a spatial coordinate in a metric space (such that the distance is well-defined). The multivariate time series \(\mathbf{X} \in \mathbb{R}^{N \times T \times C}\) records a \(C\)-dimensional physical quantities of interest generated by \(N\) series (i.e., sensors or instances) over \(T\) time steps.  
With a given threshold $\epsilon$, the spatial coordinates induce a spatial graph \(G = (V, E)\), where each node corresponds to a series (i.e., $|V| = N$) and two nodes are connected by an edge $e \in E$ if their spatial distance is smaller than $\epsilon$. Besides, we use \(\mathbf A \in \mathbb{R}^{N\times N}\) to represent the adjacent matrix of $G$.

\textbf{Spatiotemporal Series Forecasting}.
Given the past \(T\) steps historical observations \(\mathbf X_{t-T+1:t}\) and the spatial network \(G\), the goal of the spatial-temporal series forecasting is to predict the future \(F\) steps of spatial-temporal series \(\hat{\mathbf X}_{t+1:t+F}\), 
\begin{equation}
\hat{\mathbf X}_{t + 1:t+F}=\mathcal{F}_{\theta}(\mathbf X_{t - T + 1:t},G)
\end{equation}
where \(\mathcal{F}_{\theta}(\cdot)\) denotes the spatiotemporal forecasting model parameterized by \(\theta\).

\section{Methodology}

\begin{figure*}
    \centering
    \includegraphics[width=0.7\textwidth]{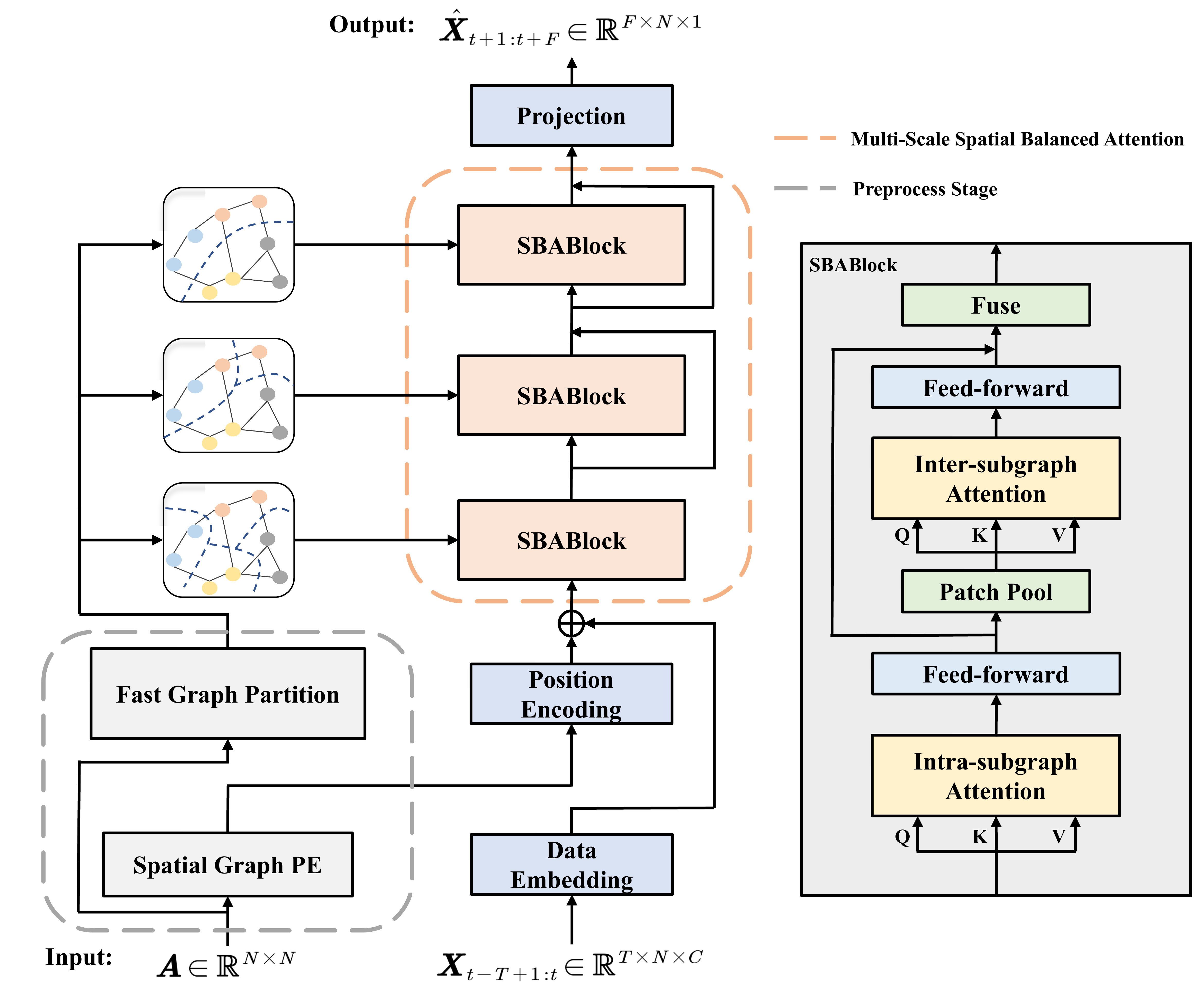}
    \vspace{-5mm}
    \caption{Left: The framework of our proposed Spatial Balance Transformer. Right: The structure of the proposed Spatial Balance Attention Block.}
    \label{fig:model}
\end{figure*} 

\textbf{Pipeline}. The framework of our proposed forecasting model dubbed Spatial Balance Transformer (SBATransformer) is shown in the left of~\cref{fig:model}. Following the setting of previous work~\cite{liu2024itransformer,cai2024forecastgrapher}, we first reshape and transform the historical observation $\mathbf X_{t-T+1:t} \in \mathbb R^{N \times T \times C}$ into the input embeddings $\mathbf X_\mathrm{emb} \in \mathbb R^{N \times D}$ with a linear map, where $D$ is the embedding dimension. Next, the embedding $\mathbf X_\mathrm{emb}$ and graph partition $\mathcal G$ (as illustrated shortly) are fed into the Spatial Balance Attention (as illustrated in the right of~\cref{fig:model}) to fuse the information and form the contextual representations, which will be used to make the forecasting (Section~\ref{sec:sba-block}). Building on the proposed blocks, we further develop a multi-scale attention mechanism to enhance representation learning by stacking multiple Spatial Balance Attention layers with various spatial scales (Section~\ref{sec:multi-scale-attention}).

\textbf{Design Motivation}. To model the spatial correlation, the existing spatiotemporal forecasting methods either simply rely on the spatial graph to represent the underlying topology~\cite{li2017diffusion,yu2017spatio,zhao2019t} or treat each node pairs equally and learn the pairwise correlation with attention in a data-driven way~\cite{wu2019graph, bai2020adaptive, wu2020connecting, huang2023crossgnn}. The former approaches excel in capturing local correlation but struggle to exchange information for distant node pairs. The latter ones are able to discover globally correlated pairs but fail to exploit spatial proximity prior, and consequently, the learned attention maps are often unstructured (lack spatial sparsity) and the computational costs are expensive. To address the dilemma, we need to carefully balance the modeling of global and local correlation as well as computational efficiency. Our key observation is that the spatial graph often exhibits stronger correlations within local areas, which is also supported by the first law of geography~\cite{miller2004tobler}. 

This motivates us to partition the spatial graph $G$ into a collection of subgraphs $\mathcal G$ and we can then fuse the information locally in each subgraph with self-attention, that is, Intra-Subgraph Attention; next, we treat each subgraph as a node and then exchange information via self-attention again, that is, Inter-subgraph Attention. As both steps are differentiable, the message from distant nodes can be backpropagated through Inter-subgraph Attention to the relevant nodes in a parsimonious fashion. Such a two-stage attention mechanism enables us to effectively capture both the local and global correlations, and meanwhile reducing the computational burdens significantly. To partition the graph, we opt for the METIS algorithm~\cite{karypis1998fast} given its efficiency and balanced subgraph outputs.      

\subsection{Spatial Balanced Attention Block}
\label{sec:sba-block}
In this section, we introduce details of the Spatial Balance Attention Block (SBABlock), guided by Tobler’s first and second laws of geography. The block comprises an Intra-subgraph Attention responsible for fine-grained correlation modeling within subgraphs and an Inter-subgraph Attention accounting for coarse-grained correlation modeling across subgraphs. Firstly, we reshape the input \(\mathbf{X}_\mathrm{emb}\) from \(\mathbb{R}^{N \times D}\) to \(\mathbb{R}^{P \times M \times D}\) according to the graph partition result $\mathcal G$, where \(P\) is the number of subgraphs, and \(M\) is the number of nodes in the largest subgraph. If the number of nodes in the subgraph is less than \(M\), we pad it with zeros and mask it in subsequent attention operations.

\textbf{Intra-subgraph Attention}. In light of Tobler’s first law of geography~\cite{miller2004tobler}, we first employ intra-subgraph attention to model the local spatial correlation. Formally, let \(\mathbf{X}\) be the input of the block (e.g., $\mathbf{X}_\mathrm{emb}$) and \(\mathbf{X}_{p} \in \mathbb{R}^{M \times D}\) denote the embedding in the $p$-th subgraph, the representations within the subgraph are updated as follows.
\begin{align}
    \boldsymbol{\alpha}_p &= \mathrm{softmax}\left( \frac{\mathbf{Q}_{p} \mathbf{K}_{p}^{\top}} {\sqrt{d \,}} \right) \mathbf{V}_{p} \\
    \mathbf{Y}_{p} &= \mathrm{FFN}\left( \boldsymbol{\alpha}_p \right)
\end{align}
\begin{equation}
    \mathbf{Q}_p, \mathbf{K}_p, \mathbf{V}_p = \left(\mathbf{W}_{Q},\mathbf{W}_{K},\mathbf{W}_{V}\right) \mathbf{X}_p
\end{equation}
where \(\mathbf{W}_{Q}\), \(\mathbf{W}_{K}\) and \(\mathbf{W}_{V}\) are learnable parameters that transform $\mathbf X_p$ into different semantic spaces, \(\boldsymbol{\alpha}_p \in \mathbb{R}^{M \times M}\) is the intra-subgraph attention map that captures the local spatial correlation in the $p$-th subgraph, and $\mathbf Y_p$ is the learned contextual representations that will be fed to the subsequent inter-subgraph attention.

\textbf{Inter-subgraph Attention}. Tobler’s second law of geography \cite{miller2004tobler} states that the phenomenon external to a geographic area of interest affects what goes on inside. However, as mentioned in Section~\ref{sec:intro}, directly using the attention mechanism will lead to quadratic complexity $\mathcal O(N^2)$ and incur extra noise that impairs the forecasting performance. In contrast to the existing methods~\cite{liu2024itransformer}, we innovatively learn the attention between different subgraphs to approximate the global attention mechanism. To this end, we first apply mean pooling to \(\mathbf{Y}_p\) to obtain the subgraph representation \(\mathbf{s}_{p} \in \mathbb{R}^{D}\) and then stack them to produce \( \mathbf{S} \triangleq [\mathbf s_1; \mathbf s_2; \ldots ;\mathbf s_P] \in \mathbb{R}^{P \times D}\). Then, we employ the inter-subgraph attention to exchange information across subgraphs as follows.
\begin{align}
    \boldsymbol{\alpha}^\prime &= \mathrm{softmax}\left( \frac{\mathbf{Q}^\prime \mathbf{K}^{\prime \top}} {\sqrt{d \,}} \right) \mathbf{V}^\prime \\
    \mathbf{S}^\prime &= \mathrm{FFN}\left( \boldsymbol{\alpha}^\prime \right) \label{eq:s-prime}
\end{align}
\begin{equation}
    \mathbf{Q}^\prime, \mathbf{K}^\prime, \mathbf{V}^\prime = \left(\mathbf{W}_{Q^\prime},\mathbf{W}_{K^\prime},\mathbf{W}_{V^\prime}\right) \mathbf{S}
\end{equation}
where \(\mathbf{W}_{Q'}\), \(\mathbf{W}_{K'}\), and \(\mathbf{W}_{V'}\) are learnable parameters. \(\boldsymbol{\alpha}^\prime \in \mathbb{R}^{P \times P}\) captures the global spatial correlation among subgraphs, which allows us to approximate the global attention mechanism by assigning the same attention value $\alpha_{pq}$ to any pair of nodes from the subgraphs $p$ and $q$. This can be considered as a sort of \emph{implicit regularization} that encourages the model to focus on mining groups of nodes that share similar patterns while ignoring the unrelated noise. In other words, we only permit the distant node pairs to exchange information in a parsimonious manner, and only valuable information can be transited across groups. 

In the end, to take both local and global spatial information into account, we expand the shape of \(\mathbf{S}^{\prime}\) in Eq.~\ref{eq:s-prime} from \(\mathbb{R}^{P \times D}\) to \(\mathbb{R}^{P \times M \times D}\), which is concatenated with the local representation and then transformed by a linear map to produce the output:
\begin{equation}
     \mathbf{{X}}^\prime = \mathbf{W} (\mathbf{Y} \| \mathbf {S}^\prime)
\end{equation} 
where \(\|\) indicates the concatenation operator and \(\mathbf{W} \in \mathbb{R}^{2D \times D}\) is the parameter of linear map layer. 

\textbf{Remark}. By a combination of intra-subgraph and inter-subgraph attention, our proposed Spatial Balance Attention block naturally incorporates spatial proximity as a soft inductive bias into the model design while also offering sufficient flexibility to enable efficient long-distance information exchange. Guided by the first and the second law of geography, it strikes a balance between exploiting spatial proximity and capturing global correlation. The resulting forecasting model can therefore produce structured spatial correlation and scale to large real-world spatiotemporal datasets. In addition, as only the attention mechanism is involved, our proposed block is easy to implement and exhibits good generalizability across various datasets.

\subsection{Multiscale Spatial Balance Architecture}
\label{sec:multi-scale-attention}
In practice, the spatial correlation often present multiscale structures due to the multiscale property of underlying physical dynamics. To capture the intrinsic multiscale property, we develop a multiscale spatial balance architecture by stacking $L$ SBA-Blocks by gradually increasing the subgraph scales. Specifically, we perform graph partition with various subgraph scales to produce $L$ sets, $\mathcal G_1, \mathcal G_2, \ldots, \mathcal G_L$ such that $|\mathcal G_i| = |\mathcal G_{i-1}| / 2$. $L$ is set to $3$ by default and we empirically find that it performs quite well in practice. such a design also brings two additional benefits: 1) it progressively expands the receptive field of the node attention mechanism, which aligns with the spatial diffusion process of many spatiotemporal physical dynamics; 2) it offers the chance for spatially-closed nodes at the boundary of two subgraphs to exchange information in the high-level block.


\textbf{Position Encodings}. Self-attention serves as the core computational module in our proposed block, however, it is oblivious to the local graph structures due to its permutation invariant property. To encode the structural information into it, we adopt the graph Laplacian eigenvectors as the position encoding. Following the suggestion of prior work~\cite{DBLP:conf/nips/KreuzerBHLT21}, the eigenvectors of top-$k$ smallest eigenvalues are chosen as the node position encoding. The eigenvectors of the graph Laplacian can be obtained by the Singular Value Decomposition very cheaply for the small matrices, e.g., $2000 \times 2000$. To accelerate the computation of large matrices, we propose to exploit the sparsity of spatial graphs again by partitioning the graph into small subgraphs (with size $2000 \times 2000$) and computing the eigenvectors within each subgraph.

\textbf{Forecasting}. We reshape the output of \(L\)-th block \( \mathbf{X}^{\prime(L)} \) from \(\mathbb{R}^{P \times M \times D}\) to \(\mathbb{R}^{N \times D} \) according to the graph partition result and produce the multi-step prediction \(\hat{\mathbf X}_{t_0:t_0 + F}\) through a linear projection. The model is optimized by minimizing mean absolute error:
\begin{equation*}
\mathcal{L}(\mathbf X_{t_0:t_0 + F}, \hat{\mathbf X}_{t_0:t_0 + F}) = \frac{\sum_{n = 1}^{N} \sum_{t = t_0}^{t_0 + F - 1} |\hat{x}_t^{(n)} - x_t^{(n)}|}{N \times F}
\end{equation*}

\subsection{Complexity Analysis}
Since graph partition and graph Laplacian eigenvector calculation can be rapidly completed in the preprocessing stage, our method achieves significant computational efficiency improvements by focusing on intra-subgraph and inter-subgraph attention. Let $M, P$, and $D$ denote the largest subgraph size, the number of subgraphs, and the dimension of hidden representation, respectively. The computational cost for our proposed Spatial Balance Attention is $\mathcal{O}(M^2D + P^2 D)$, which has much better scalability than the methods of quadratic complexity. Note that the efficiency of our model can be further enhanced through linear attention techniques.
We will empirically evaluate the efficiency of different methods in Section~\ref{sec:efficiency-study}.

\section{Experiments}

\begin{table}
\caption{Dataset statistics.}
\vspace{+4mm}
\label{table:dataset-table}
\begin{center}
\begin{small}
\begin{tabular}{ccccc}
\toprule
Source                                       & Timesteps              & Frequency                  & Dataset   & Nodes \\ \midrule
\multicolumn{1}{c}{\multirow{4}{*}{LargeST}} & \multirow{4}{*}{35040} & \multirow{4}{*}{15 Minute} & SD        & 716   \\
\multicolumn{1}{c}{}                         &                        &                            & GBA       & 2352  \\
\multicolumn{1}{c}{}                         &                        &                            & GLA       & 3834  \\
\multicolumn{1}{c}{}                         &                        &                            & CA        & 8600  \\ \midrule
\multirow{3}{*}{PV-US}                       & \multirow{3}{*}{52560} & \multirow{3}{*}{10 Minute}  & WEST & 1082  \\
                                             &                        &                            & EAST & 4084  \\
                                             &                        &                            & ALL  & 5016  \\ \bottomrule
\end{tabular}
\end{small}
\end{center}
\end{table}

\begin{table*}[]
\renewcommand{\arraystretch}{1.5}
\caption{Spatiotemporal series forecasting performance comparison. The best results are highlighted in \textbf{bold}, while the second-best results are \uline{underlined}. - indicates out of memory.} 
\label{table:performance-comparison}
\begin{center}
\begin{small}
\resizebox{\linewidth}{!}{
\begin{tabular}{l|cc|cc|cc|cc|cc|cc|cc}
\toprule
Dataset      & \multicolumn{2}{c|}{SD}         & \multicolumn{2}{c|}{GBA}        & \multicolumn{2}{c|}{GLA}        & \multicolumn{2}{c|}{CA}         & \multicolumn{2}{c|}{WEST}     & \multicolumn{2}{c|}{EAST}     & \multicolumn{2}{c}{ALL}       \\ \midrule
Metric       & MAE            & RMSE           & MAE            & RMSE           & MAE            & RMSE           & MAE            & RMSE           & MAE           & RMSE          & MAE           & RMSE          & MAE           & RMSE          \\ \midrule
PatchTST     & 18.95          & 31.17          & 21.08          & 34.98          & 20.11          & 33.16          & 19.77          & 32.80          & 3.64          & 7.91          & 5.78          & 9.44          & 4.17          & 7.70          \\
TimeMixer    & 18.59          & 30.13          & 20.21          & 33.07          & 20.34          & 33.08          & 18.78          & 30.81          & 3.94          & 7.98          & 3.41          & 6.61          & 3.52          & 6.87          \\
STGCN        & 18.38          & 32.81          & 22.76          & 39.32          & 20.91          & 37.43          & 20.45          & 37.08          & 3.96          & 7.96          & 3.30          & 6.52          & 3.49          & 6.85          \\
MTGNN        & 17.13          & 29.03          & 19.46          & 32.07          & 19.67          & 32.11          & 17.67          & 30.03          & 3.71          & 7.88          & 3.25          & 6.57          & 3.37          & 6.92          \\
CrossGNN     & 21.85          & 35.51          & 22.75          & 36.57          & 22.93          & 36.44          & 21.20          & 34.53          & 4.00          & 8.21          & 3.57          & 6.96          & 3.63          & 7.24          \\
iTransformer & 17.60          & 29.41          & 19.07          & 32.12          & 18.31          & 30.47          & -              & -              & 3.68          & 7.95          & 3.16          & 6.45          & 3.66          & 7.30          \\
Crossformer   & 16.94          & 28.36          & {\ul 18.85}    & {\ul 31.02}    & 18.64          & 31.11          & {\ul 16.98}    & {\ul 28.34}    & {\ul 3.54}    & {\ul 7.85}    & {\ul 2.98}    & {\ul 6.22}    & {\ul 3.08}    & {\ul 6.55}    \\
Card         & 18.84          & 30.93          & 20.56          & 33.74          & 20.14          & 32.88          & 18.80          & 31.18          & 3.69          & 7.93          & 3.32          & 6.63          & 3.37          & 6.93          \\
RPMixer      & 17.65          & 29.08          & 19.60          & 31.94          & 19.92          & 32.53          & 18.28          & 30.07          & 3.89          & 7.91          & 3.43          & 6.57          & 3.55          & 6.85          \\
BigST        & 20.09          & 33.20          & 20.96          & 34.23          & 21.73          & 35.69          & 20.96          & 34.69          & 3.72          & 7.93          & 3.25          & 6.59          & 3.33          & 6.81          \\
Sumba        & {\ul 16.49}    & {\ul 27.47}    & 18.87          & 31.39          & {\ul 18.30}    & {\ul 30.23}    & 17.02          & 28.42          & 3.63          & 7.68          & 3.15          & 6.33          & 3.29          & 6.73          \\
SOFTS        & 17.31          & 28.83          & 19.13          & 32.41          & 18.52          & 31.02          & 17.55          & 29.62          & 3.77          & 8.28          & 3.18          & 6.56          & 3.27          & 6.86          \\ \midrule
SBATransformer    & \textbf{15.95} & \textbf{26.93} & \textbf{17.44} & \textbf{29.86} & \textbf{17.07} & \textbf{28.71} & \textbf{15.67} & \textbf{26.88} & \textbf{3.44} & \textbf{7.53} & \textbf{2.81} & \textbf{5.84} & \textbf{2.88} & \textbf{6.17} \\ \bottomrule
\end{tabular}}
\end{small}
\end{center}
\end{table*}

In this section, we evaluate our approach on seven benchmark datasets (Section~\ref{sec:performance-comparison} and Appendix~\ref{sec:full-experiments-results}). Section~\ref{sec:efficiency-study} presents the efficiency analysis and Section~\ref{sec:ablation-study} describes the ablation study. The sensitivity of hyperparameters is detailed in Section~\ref{sec:parameter-sensitivity-analysis} and Appendix~\ref{sec:more-hyperparameters-sensitivity}. To gain a more profound understanding of our model, we also conducted visualizations on the SD and GBA datasets, which are included in Appendix~\ref{sec:visualization}. 

\subsection{Experimental Setup}

\textbf{Datasets.} 
We evaluate the performance and efficiency of the proposed method on large-scale spatiotemporal series datasets by taking novel traffic and energy analytics benchmarks into account. The first benchmark is the LargeST datasets \cite{liu2024largest}, which contain traffic data collected from 8,600 sensors in California in 2019. The second benchmark is the PV-US datasets \cite{hummon2012sub}, consisting of energy production by 5016 PV farms scattered across the United States in 2006. Dataset statistics are presented in Table~\ref{table:dataset-table}, and further particulars are available in Appendix~\ref{sec:implementation-details}.

\textbf{Baselines.}
We compare our method with the following baselines: (1) no spatial correlation modeling methods: {PatchTST}~\cite{patchtst2023}, TimeMixer~ \cite{wang2024timemixer}; (2) Local spatial correlation modeling methods: STGCN~ \cite{yu2017spatio}; (3) Global spatial correlation modeling methods: MTGNN~ \cite{wu2020connecting}, CrossGNN~\cite{huang2023crossgnn}, iTransformer~ \cite{liu2024itransformer}; (4) Scalable spatial correlation modeling methods: Crossformer~\cite{zhang2023crossformer}, Card~\cite{wang2024card}, RPMixer \cite{yeh2024rpmixer}, BigST~\cite{han2024bigst}, Sumba~\cite{chenstructured}, SOFTS~\cite{han2024softs}. More details of baselines are provided in Appendix~\ref{sec:implementation-details}.

\textbf{Evaluation Metrics.}
We conduct a comprehensive comparison using various evaluation criteria from the performance and efficiency perspectives. We evaluated performance using the mean absolute error (MAE), root mean square error (RMSE), and mean absolute percentage error (MAPE). We consider efficiency by measuring both the training wall-clock time and maximum memory usage during training. 

\textbf{Implementation details.}
 We set the look-back time window \(T\) to 96 and 144 for LargeST and PV-US datasets, aiming to predict the observations for the next 12 steps. The dimensions of the input embedding and hidden embedding \(D\) are set to 512. The number of blocks \(L\) is set to 3. The initial number of subgraphs \(P\) for CA, ALL, EAST, GLA, GBA, WEST, and SD is set to 128, 64, 8, 64, 8, 16, and 8. The epsilon \(\epsilon\) is set by following the suggestion of DCRNN \cite{li2017diffusion}. All experiments in this study are implemented using PyTorch~\cite{paszke2019pytorch} and conducted on NVIDIA RTX 4090 GPU with 24GB memory. We run each experiment three times and report the average results.

\begin{figure*} \centering   
\subfigure{    
\includegraphics[width=0.9\textwidth]{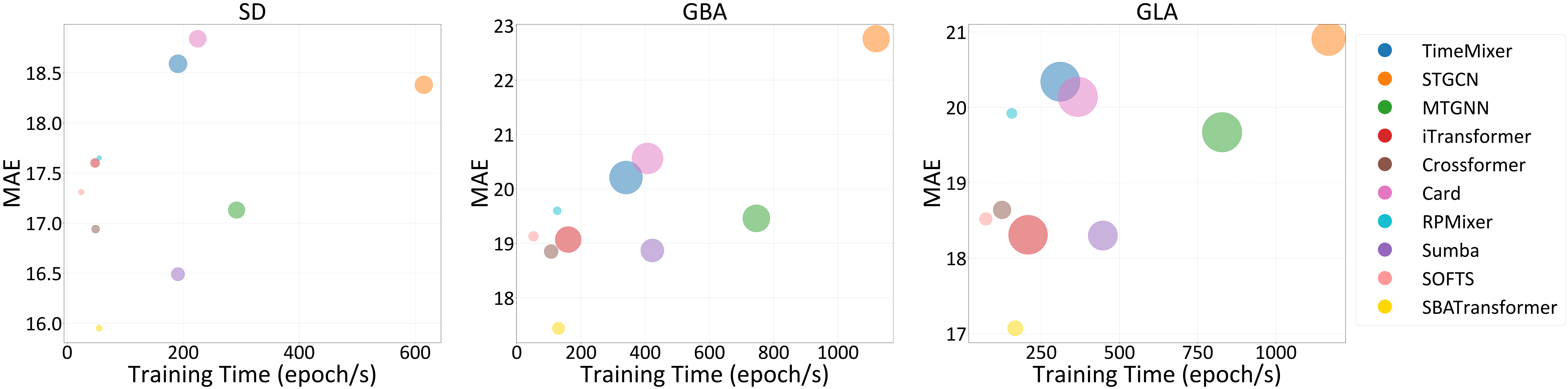}  
}    
\subfigure{    
\includegraphics[width=0.9\textwidth]{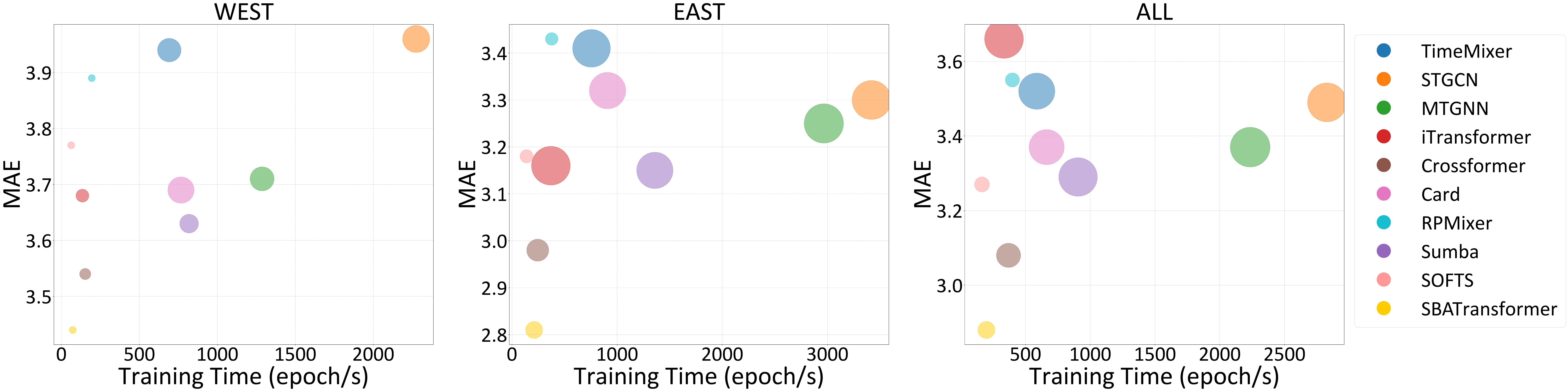}  
}     
\caption{Spatiotemporal series forecasting efficiency comparison.}     
\label{fig:efficiency-comparison-LargeST-PVUS}   
\end{figure*}
\subsection{Performance Comparison}
\label{sec:performance-comparison}
Table~\ref{table:performance-comparison} presents the average forecasting performance of the next 12 steps for all methods on seven spatiotemporal datasets. The best results are highlighted in \textbf{bold}, and the second-best results are \uline{underlined}. The symbol – indicates the out of memory error. To ensure a fair comparison, we follow the official configurations of baselines and ignore the pre-training stage for BigST. The full results are provided in Appendix~\ref{sec:full-experiments-results}. The experimental conclusions are as follows.

First, the sub-optimal performance of PatchTST and TimeMixer highlights the importance of spatial correlation modeling. Second, compared with the local spatial correlation modeling model STGCN, MTGNN improves performance with a larger receptive field. Dynamic models like iTransformer, Crossformer, Card, and Sumba perform more excellently in normal-scale datasets, highlighting the significance of dynamic spatial correlation modeling. Our method outperforms all baselines, showing up to 4.1\% improvement on the medium-scale WEST dataset due to the structured and dynamic spatial correlation modeling. Ultimately, the quadratic complexity of the above models limits their application in large-scale datasets, further highlighting the advantages of scalable spatial correlation modeling approaches. However, both linearized spatial correlation computation (BigST, Sumba) and mixing channels in spatial dimension (Crossformer, RPMixer, SOFTS) suffer from information loss due to spatial reduction. In contrast, our method achieves state-of-the-art performance, demonstrating an improvement of up to 7.7\% on the large-scale dataset CA. This superior performance is attributed to the structured spatial correlation constraints, namely accurate local spatial correlation and restricted global spatial correlation, which preserve key information and filter out noise.

\subsection{Efficiency Study}
\label{sec:efficiency-study}
To evaluate model efficiency, we present the training time consumption and maximum memory usage during the training of our method and previous models. We utilize \textit{torch.nn.DataParallel} to train models across multiple GPUs. When measuring computational efficiency, we fix the number of available GPUs for different datasets and maximize the batch size up to 64. Specifically, the number of available GPUs for CA, ALL, EAST, GLA, GBA, WEST, and SD is set to 8, 8, 4, 4, 2, 1, and 1. When measuring memory consumption, we use as many GPUs as possible to reach a batch size of 64, with a maximum of 8 GPUs used. We intuitively compare the model efficiency by plotting a bubble chart for each dataset as shown in Figure~\ref{fig:efficiency-comparison-LargeST-PVUS}, in which the horizontal axis represents the training time per epoch (in seconds), the vertical axis represents the model performance (in MAE), and the size of the bubble corresponds to the maximum memory used by the model during training. The bubble chart for the CA dataset and detailed numerical comparison are provided in Appendix~\ref{sec:full-experiments-results}. The experimental conclusions are as follows.

First, even without spatial correlation modeling, when the number of nodes and correspondingly the number of samples increases, the memory consumption of TimeMixer is also not optimistic. Second, existing spatial correlation modeling models (STGCN, MTGNN, iTransformer) experience rapid growth in computational and/or memory consumption as the number of nodes increases due to quadratic complexity. Some existing scalable models also face this problem due to the presence of complex time dependency modeling module (Card) or dynamic spatial correlation modeling module (Sumba). Last, the channel mixing approach (Crossformer, RPMixer, SOFTS), which also has relatively low computational and memory consumption, can only achieve suboptimal performance. In contrast, our model confines fine-grained spatial correlation modeling within local subgraphs and enables efficient long-distance information exchange through spatial correlation modeling between subgraphs. This not only filters out noise to improve performance but also effectively reduces memory consumption. In particular, it achieves up to 4× and 5× improvements on large-scale GLA and ALL datasets compared with previous global spatial correlation modeling models.

\subsection{Ablation Study}
\label{sec:ablation-study}
\begin{table}[]
\caption{Ablation study of the GLA and CA datasets. The best results are highlighted in \textbf{bold}.} 
\label{table:ablation-study-GLA-CA}
\begin{center}
\begin{small}
\resizebox{\linewidth}{!}{
\begin{tabular}{l|ccc|ccc}
\toprule
Dataset        & \multicolumn{3}{c|}{GLA}                         & \multicolumn{3}{c}{CA}                             \\ \midrule
Metric         & MAE            & RMSE           & MAPE(\%)       & MAE            & RMSE           & MAPE(\%)         \\ \midrule
w/o. PE        & 17.23          & 29.03          & 10.62          & 16.10          & 27.42          & 11.60            \\
w/o. Intra-Att & 18.52          & 30.46          & 11.94          & 18.07          & 30.27          & 13.79            \\
w/o. Inter-Att & 17.33          & 29.13          & 10.62          & 16.01          & 27.30          & 11.84            \\
w/o. MA        & 17.21          & 28.93          & 10.83          & 16.19          & 27.49          & 11.56            \\ \midrule
SBATransformer      & \textbf{17.07} & \textbf{28.71} & \textbf{10.36} & \textbf{15.67} & \textbf{26.88} & \textbf{11.15\%} \\ \bottomrule
\end{tabular}}
\end{small}
\end{center}
\end{table}
We conduct the ablation study on the LargeST datasets to validate the effectiveness of the proposed modules. Specifically, we consider the following variants of our proposed model:
(1)\textbf{ w/o. PE}: We remove the graph position encodings in the preprocessing stage. (2) \textbf{w/o. Inter-Att}: We remove the inter-subgraph attention and only capture local spatial correlation. (3) \textbf{w/o. Intra-Att}: We remove the intra-subgraph attention and only capture the spatial correlation between subgraphs. (4) \textbf{w/o. MA}: We remove the multiscale architecture and perform graph partitions with the same subgraph scales. As shown in Table~\ref{table:ablation-study-GLA-CA} and Table~\ref{table:ablation-study-SD-GBA}, key observations are as follows.

First, the decline in the performance of \textbf{w/o.\,PE} verifies the necessity of the graph positional encoding, which reveals the structural information of the graph that is otherwise lost in the vanilla Transformer. Second, the performance of \textbf{w/o.\,Intra-Att} declines rapidly by up to 13.3\%, 11.2\%, and 19.1\% on three metrics. This indicates that local spatial correlation is crucial in accurate large-scale spatiotemporal series forecasting, which aligns with Tobler’s first law of geography. Meanwhile, the performance of \textbf{w/o.\,Inter-Att} drops by up to 5.8\%, suggesting that global spatial correlation is also helpful for exact forecasting, in accordance with Tobler’s second law of geography. In the end, the performance of \textbf{w/o.\,MA} drops by up to 3.2\%, which proves the effectiveness of the proposed multiscale architecture.

\subsection{Parameter Sensitivity Analysis}
\label{sec:parameter-sensitivity-analysis}
\begin{figure}
    \centering
    \includegraphics[width=0.48\textwidth]{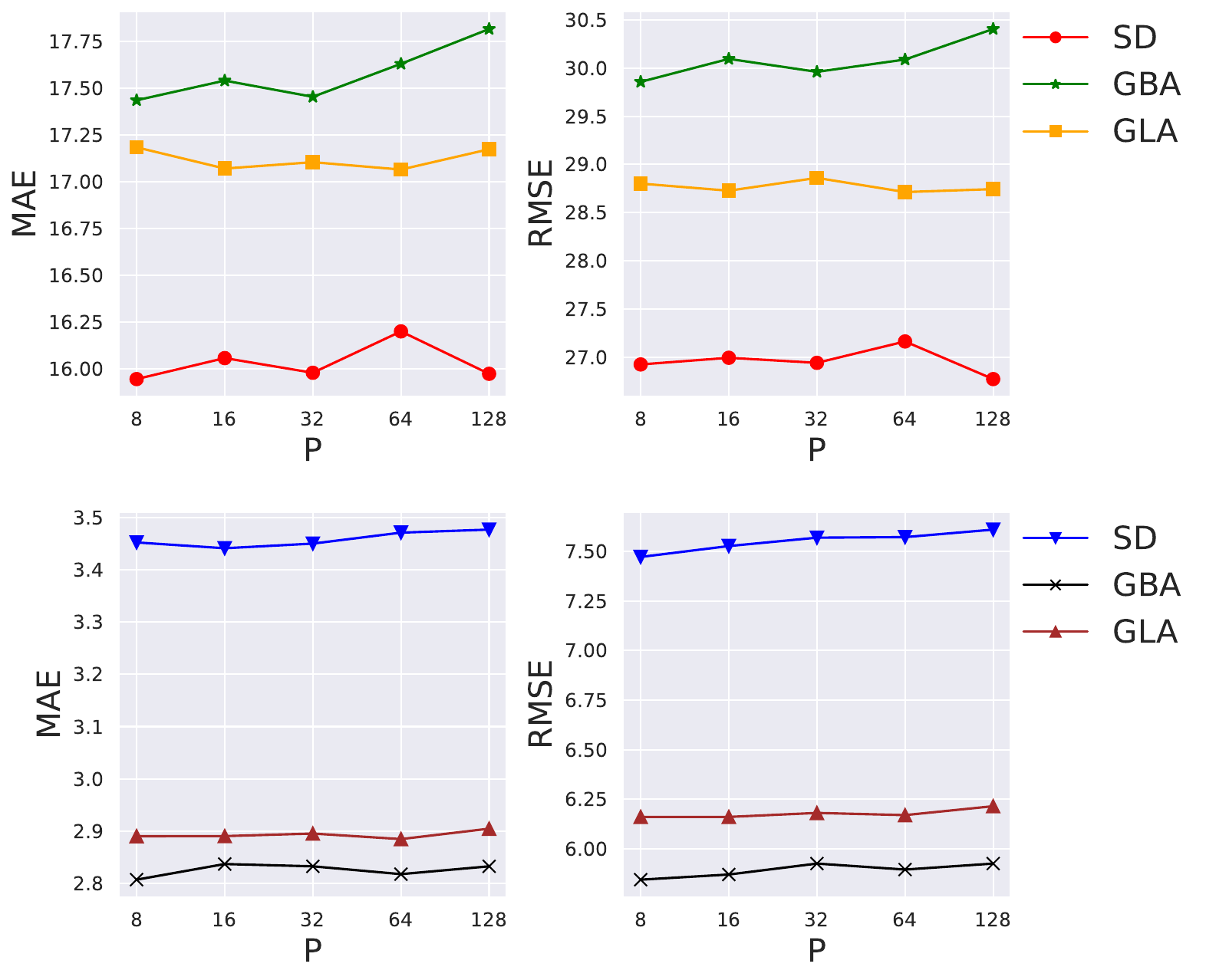}
    \vspace{-5mm}
    \caption{The sensitivity of our method to the initial number of subgraphs \(P\).}
    \label{fig:hyperparamter-sensitivity-M}
\end{figure}
We evaluate the impact of the initial number of subgraphs \(P\) on the performance of our model, where it ranges from 8, 16, 32, 64 to 128. A larger value implies fewer nodes within a subgraph, resulting in a more rigorous modeling of local spatial correlation. Conversely, a smaller value means more nodes in a subgraph, which broadens the scope of local space, but leads to higher memory and computational consumption. As shown in Figure~\ref{fig:hyperparamter-sensitivity-M}, the experimental results demonstrate that our model has a low sensitivity to the parameter \(P\). This implies that an overly large receptive field has limited utility for accurate spatiotemporal series forecasting when modeling spatial correlation, which is consistent with our proposition. The sensitivity analysis of look-back time window \(T\) and the number of blocks \(L\) is presented in Appendix~\ref{sec:more-hyperparameters-sensitivity}. 
\section{Conclusion}
In this paper, we propose a novel Spatial Balanced Attention block
for spatiotemporal forecasting that strikes a balance between obeying spatial proximity and capturing global correlation. Specifically, we partition the spatial graph into a set of subgraphs and utilize the Intra-subgraph and Inter-subgraph Attention to learn local and global spatial correlation. Building on the proposed block, we develop a multiscale spatiotemporal forecasting model by progressively increasing the subgraph scales. The resulting model is both scalable and able to produce structured spatial correlation. Performance comparison and efficiency analysis on seven real-world spatiotemporal datasets validate the superiority of our method in medium and large-scale scenarios. Extensive ablation studies confirm the effectiveness of each proposed component. In the future, we would like to explore how to extend our method to general multivariate time series data. Moreover, we aim to explore the integration of our proposed spatial correlation module with other efficient temporal dependency encoders, developing efficient pre-trained models for general spatiotemporal data analysis.
\section*{Impact Statement}
In this paper, we propose a novel scalable spatiotemporal forecasting model to capture structured spatial correlation guided by Tobler's law of geography. Our research aims to contribute positively to the relevant community while ensuring no negative social impact.
\bibliography{example_paper}
\bibliographystyle{plainnat}

\newpage
\appendix
\onecolumn
\section{Implementation Details}
\label{sec:implementation-details}
\subsection{Datasets}
All datasets used in our study are open-source or freely available for research purposes.

\textbf{LargeST.}
The LargeST datasets \cite{liu2024largest} contain traffic data collected from 8,600 sensors in California by the California Department of Transportation (CalTrans) Performance Measurement System (PeMS) \footnote{https://pems.dot.ca.gov}. It can be divided into four subsets, namely SD, GBA, GLA, and CA. These subsets respectively include sensor data from the San Diego region, the Greater Bay Area region, the Greater Los Angeles region, and all 8,600 sensors across California, spanning from 2017 to 2020. The sensor readings, which are originally recorded every 5 minutes, are aggregated into 15-minute intervals, resulting in 96 records per day. In accordance with the experimental setup, we only use the traffic data in 2019. Moreover, we directly use the spatial adjacency matrix provided by these datasets.

\textbf{PV-US.}
The PV-US\footnote{https://www.nrel.gov/grid/solar-power-data.html} datasets \cite{hummon2012sub} compile simulated energy production from 5,016 PV farms across the United States in 2006. It can be divided into three subsets: 1,082 PV plants in the West, 4,084 in the East, and 5,016 across both regions, as some plants are counted in both the western and eastern subsets. A smaller subset, often referred to as "Solar Energy" \footnote{https://github.com/laiguokun/multivariate-time-series-data}, includes data from only 137 PV plants located in Alabama and has been utilized for multivariate time series forecasting \cite{lai2018modeling}. Since the model does not have access to weather information, PV production at neighboring farms is instrumental in obtaining good predictions. We receive the preprocessed datasets from the SGP \cite{cini2023scalable}, in which data points are generated every 10 minutes, resulting in 144 records per day. To get a spatial adjacency matrix, we calculate the pairwise distances between plants based on their longitude and latitude coordinates and construct the adjacency matrix using a threshold Gaussian kernel.

Notably, the LargeST and PV-US datasets are at least an order of magnitude larger than the datasets typically used for benchmarking spatial-temporal series forecasting models. We chronologically split the datasets into training, validation, and test datasets with the ratio 0.6/0.2/0.2 and apply Z-score normalization to normalize the input data for model training.
\subsection{Baselines}
We compare the proposed approach with the following advanced baselines:
\begin{itemize}
    \item PatchTST: It introduces a patch-based attention mechanism and channel-independent strategy to establish long-term temporal dependency.
    \item TimeMixer: It proposes Past-Decomposable-Mixing and Future-Multipredictor-Mixing blocks to take full advantage of disentangled multiscale series in both past extraction and future prediction phases.
    \item STGCN: It utilizes graph convolutions to model local spatial correlation and 1D convolutions for temporal modeling in traffic forecasting.
    \item MTGNN: It builds an adaptive static global directed graph using learnable node embedding and aggregates information along spatial dimensions through mix-hop propagation.
    \item CrossGNN: It proposes a Cross-Scale GNN to extract scales with clearer trend and weaker noise, and a Cross-Variable GNN to utilize the global homogeneity and heterogeneity between different variables.
    \item iTransformer: It embeds the whole time series into a spatial token and captures dynamic global spatial correlation using the self-attention mechanism.
    \item Crossformer: It utilizes the Dimension-Segment-Wise embedding to embed the input data into a 2D vector array and proposes a two-stage attention layer to capture temporal dependency and channel correlation.
    \item Card: It introduces summarized spatial tokens to decrease the complexity of the spatial attention module and employs a token blend mechanism in the temporal attention module to extract local temporal dependency.
    \item RPMixer: It employs MLPs to model temporal dependency and integrates random projection layers to capture spatial correlation.
    \item BigST: It is a linear complexity spatial-temporal GNNs that efficiently exploit long-range spatial-temporal dependency for large-scale traffic forecasting.
    \item Sumba: It proposes a dynamic global spatial structure generation function with well-constrained output space to model the spatial correlation and utilizes multi scale TCN to capture the temporal dependency.
    \item SOFTS: It builds a novel STAR module that aggregates all series into a global core representation and then dispatches and fuses it with individual series representations for effective channel correlation modeling.
\end{itemize}
We obtain the code of baselines directly from their corresponding GitHub repositories. For the model- and training-related configurations, we follow the recommended settings provided in their code.
\section{Full Experiments Results}
\label{sec:full-experiments-results}
\begin{figure}
    \centering
    \includegraphics[width=0.7\textwidth]{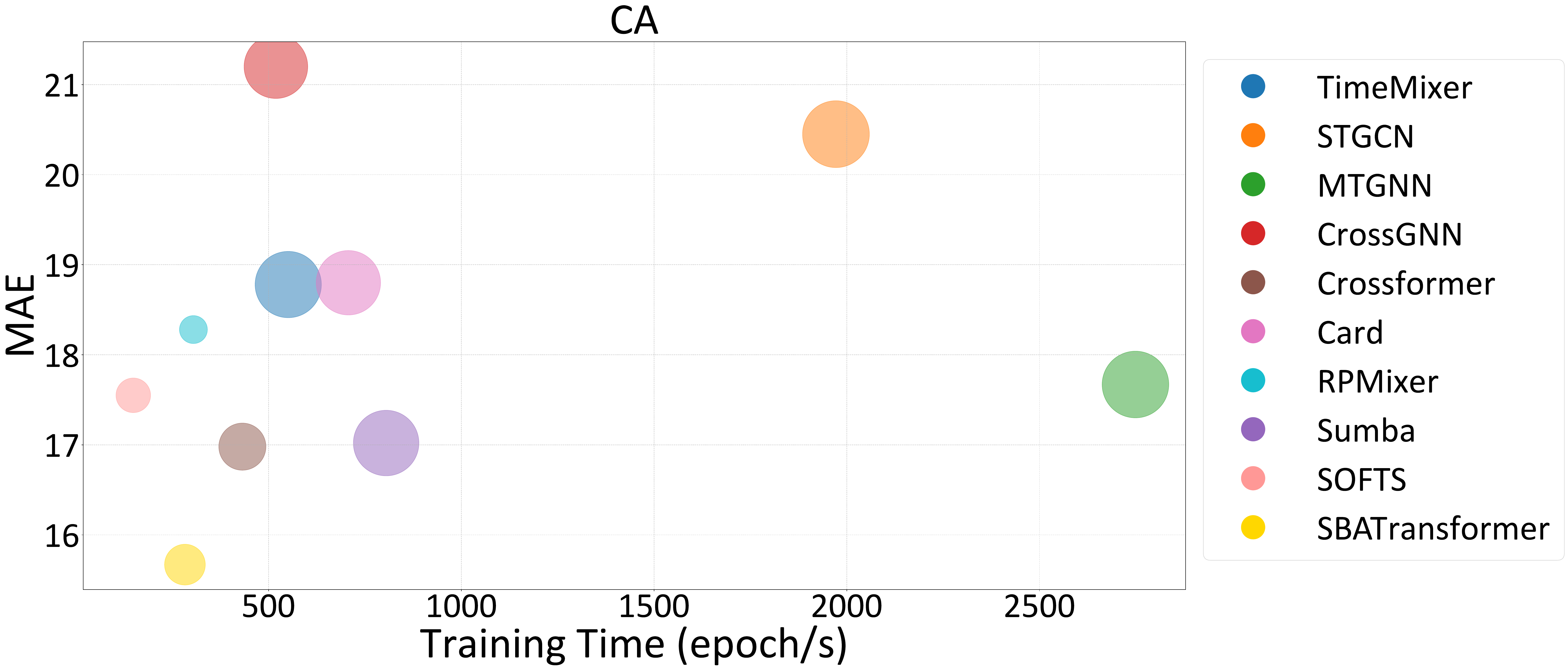}
    \vspace{-5mm}
    \caption{Efficiency comparison on the CA dataset.}
    \label{fig:effiency-analysis-CA}
\end{figure}

\begin{table}[]
\renewcommand{\arraystretch}{1.3}
\caption{Full forecasting results of the LargeST datasets. The best results are highlighted in \textbf{bold}, while the second-best results are \uline{underlined}.} 
\label{table:full-forecasting-results-of-the-LargeST-datasets}
\begin{center}
\begin{small}
\resizebox{\linewidth}{!}{
\begin{tabular}{c|c|ccc|ccc|ccc|ccc}
\toprule
\multicolumn{1}{l|}{\multirow{2}{*}{Data}} & \multirow{2}{*}{Method} & \multicolumn{3}{c|}{Horizon 3}                   & \multicolumn{3}{c|}{Horizon 6}                   & \multicolumn{3}{c|}{Horizon 12}                  & \multicolumn{3}{c}{Average}                      \\
\multicolumn{1}{l|}{}                      &                         & MAE            & RMSE           & MAPE(\%)       & MAE            & RMSE           & MAPE(\%)       & MAE            & RMSE           & MAPE(\%)       & MAE            & RMSE           & MAPE(\%)       \\ \midrule
\multirow{13}{*}{SD}                       & PatchTST                & 16.04          & 26.32          & 10.08          & 19.12          & 31.39          & 12.05          & 23.09          & 38.47          & 15.32          & 18.95          & 31.17          & 12.09          \\
                                           & TimeMixer               & 16.32          & 26.14          & 11.53          & 18.81          & 30.33          & 13.30          & 21.98          & 36.32          & 15.36          & 18.59          & 30.13          & 13.08          \\
                                           & STGCN                   & 16.92          & 29.91          & 12.02          & 18.37          & 32.61          & 12.91          & 20.59          & 37.44          & 15.08          & 18.38          & 32.81          & 13.10          \\
                                           & MTGNN                   & 15.60          & 25.93          & 12.62          & 17.52          & 29.78          & 12.91          & 19.78          & 33.95          & 14.09          & 17.13          & 29.03          & 12.59          \\
                                           & CrossGNN                & 18.33          & 29.59          & 11.48          & 22.15          & 35.71          & 14.44          & 27.30          & 35.51          & 18.01          & 21.85          & 35.51          & 14.16          \\
                                           & iTransformer            & 15.38          & 25.62          & 9.73           & 17.85          & 29.92          & 11.29          & 20.67          & 34.79          & 13.13          & 17.60          & 29.41          & 11.18          \\
                                           & Crossformer             & 15.12          & 25.31          & 10.16          & 17.10          & 28.52          & 11.49          & 19.67          & 32.95          & 13.21          & 16.94          & 28.36          & 11.47          \\
                                           & Card                    & 16.56          & 26.97          & 10.72          & 19.05          & 31.36          & 12.29          & 22.19          & 36.79          & 14.48          & 18.84          & 30.93          & 12.25          \\
                                           & RPMixer                 & 16.16          & 26.02          & 11.51          & 17.80          & 29.24          & 12.53          & 20.03          & 34.02          & 14.64          & 17.65          & 29.08          & 12.54          \\
                                           & BigST                   & 17.94          & 28.85          & 12.07          & 20.34          & 33.34          & 13.86          & 23.46          & 40.04          & 16.90          & 20.09          & 33.20          & 14.00          \\
                                           & Sumba                   & {\ul 14.70}    & {\ul 24.39}    & \textbf{9.37}  & {\ul 16.64}    & {\ul 27.72}    & {\ul 10.73}    & {\ul 19.22}    & {\ul 32.38}    & {\ul 12.88}    & {\ul 16.49}    & {\ul 27.47}    & \textbf{10.75} \\
                                           & SOFTS                   & 15.16          & 25.22          & 9.47           & 17.36          & 29.01          & 10.90          & 20.43          & 34.22          & 13.07          & 17.31          & 28.83          & {\ul 10.94}    \\
                                           & SBATransformer               & \textbf{14.35} & \textbf{23.88} & {\ul 9.42}     & \textbf{16.12} & \textbf{27.21} & \textbf{10.64} & \textbf{18.37} & \textbf{31.71} & \textbf{12.61} & \textbf{15.95} & \textbf{26.93} & \textbf{10.75} \\ \midrule
\multirow{13}{*}{GBA}                      & PatchTST                & 18.00          & 30.03          & 13.83          & 21.35          & 35.27          & 17.25          & 25.53          & 42.63          & 21.54          & 21.08          & 34.98          & 16.95          \\
                                           & TimeMixer               & 17.77          & 28.94          & 15.21          & 20.38          & 33.25          & 17.54          & 23.75          & 39.44          & 20.91          & 20.21          & 33.07          & 17.69          \\
                                           & STGCN                   & 21.24          & 36.21          & 16.50          & 23.06          & 39.86          & 18.06          & 24.87          & 43.53          & 20.25          & 22.76          & 39.32          & 18.02          \\
                                           & MTGNN                   & 17.46          & 29.06          & 13.86          & 19.67          & 32.51          & 15.66          & 22.42          & 36.68          & 18.48          & 19.46          & 32.07          & 15.73          \\
                                           & CrossGNN                & 19.41          & 31.43          & 15.26          & 23.18          & 37.03          & 18.89          & 27.69          & 44.45          & 22.99          & 22.75          & 36.57          & 18.45          \\
                                           & iTransformer            & 16.83          & 28.57          & 12.41          & 19.34          & 32.56          & {\ul 14.44}    & 22.27          & 37.39          & {\ul 16.45}    & 19.07          & 32.12          & {\ul 14.10}    \\
                                           & Crossformer             & 17.05          & 28.29          & 12.70          & {\ul 19.15}    & {\ul 31.44}    & 14.87          & 21.59          & 35.30          & 16.59          & 18.85          & {\ul 31.02}    & 14.26          \\
                                           & Card                    & 18.17          & 29.61          & 14.72          & 20.90          & 34.11          & 17.07          & 23.96          & 39.91          & 18.79          & 20.56          & 33.74          & 16.51          \\
                                           & RPMixer                 & 18.13          & 29.40          & 14.51          & 19.89          & 32.39          & 16.05          & 21.88          & 35.97          & 18.20          & 19.60          & 31.94          & 15.85          \\
                                           & BigST                   & 18.72          & 30.34          & 15.40          & 21.34          & 34.74          & 18.36          & 24.31          & 40.02          & 21.40          & 20.96          & 34.23          & 17.94          \\
                                           & Sumba                   & 17.08          & 28.53          & 13.23          & 19.24          & 31.98          & 15.32          & {\ul 21.38}    & {\ul 35.54}    & 16.98          & {\ul 18.87}    & 31.39          & 14.87          \\
                                           & SOFTS                   & {\ul 16.76}    & {\ul 28.51}    & {\ul 12.27}    & 19.39          & 32.81          & 14.78          & 22.44          & 38.19          & 17.27          & 19.13          & 32.41          & 14.46          \\
                                           & SBATransformer               & \textbf{15.92} & \textbf{27.39} & \textbf{11.62} & \textbf{17.69} & \textbf{30.21} & \textbf{13.35} & \textbf{19.77} & \textbf{33.89} & \textbf{15.48} & \textbf{17.44} & \textbf{29.86} & \textbf{13.03} \\ \midrule
\multirow{13}{*}{GLA}                      & PatchTST                & 16.79          & 27.67          & 9.75           & 20.32          & 33.40          & 12.14          & 24.75          & 41.26          & 15.79          & 20.11          & 33.16          & 12.21          \\
                                           & TimeMixer               & 17.25          & 27.86          & 10.72          & 20.57          & 33.18          & 13.38          & 24.77          & 40.91          & 16.95          & 20.34          & 33.08          & 13.32          \\
                                           & STGCN                   & 19.05          & 33.77          & 11.95          & 21.10          & 37.63          & 13.34          & 23.49          & 42.45          & 15.82          & 20.91          & 37.43          & 13.50          \\
                                           & MTGNN                   & 17.63          & 28.30          & 14.29          & 19.96          & 32.67          & 14.41          & 22.42          & 37.09          & 15.94          & 19.67          & 32.11          & 14.76          \\
                                           & CrossGNN                & 19.28          & 30.82          & 11.75          & 23.30          & 36.82          & 15.00          & 28.30          & 44.81          & 19.04          & 22.93          & 36.44          & 14.73          \\
                                           & iTransformer            & {\ul 15.93}    & {\ul 26.46}    & 9.57           & 18.56          & 30.86          & {\ul 11.39}    & 21.65          & 36.25          & 13.66          & 18.31          & 30.47          & 11.28          \\
                                           & Crossformer             & 16.30          & 26.73          & 10.58          & 18.79          & 31.19          & 11.84          & 22.24          & 37.72          & 13.52          & 18.64          & 31.11          & 11.51          \\
                                           & Card                    & 17.40          & 28.12          & 10.65          & 20.53          & 33.29          & 12.75          & 23.95          & 39.65          & 14.93          & 20.14          & 32.88          & 12.49          \\
                                           & RPMixer                 & 17.91          & 28.60          & 11.58          & 20.30          & 32.98          & 13.24          & 22.65          & 38.00          & 15.43          & 19.92          & 32.53          & 13.14          \\
                                           & BigST                   & 18.94          & 30.44          & 12.28          & 22.15          & 36.11          & 14.64          & 25.54          & 43.06          & 18.26          & 21.73          & 35.69          & 14.61          \\
                                           & Sumba                   & 17.19          & 28.17          & 10.47          & {\ul 18.53}    & {\ul 30.62}    & 11.47          & {\ul 21.13}    & {\ul 35.19}    & {\ul 13.49}    & {\ul 18.30}    & {\ul 30.23}    & 11.40          \\
                                           & SOFTS                   & 16.09          & 26.78          & {\ul 9.38}     & 18.67          & 31.21          & 11.20          & 22.06          & 37.36          & 13.61          & 18.52          & 31.02          & {\ul 11.17}    \\
                                           & SBATransformer               & \textbf{15.19} & \textbf{25.36} & \textbf{9.00}  & \textbf{17.29} & \textbf{29.05} & \textbf{10.37} & \textbf{19.79} & \textbf{33.60} & \textbf{12.77} & \textbf{17.07} & \textbf{28.71} & \textbf{10.36} \\ \midrule
\multirow{12}{*}{CA}                       & PatchTST                & 16.18          & 26.90          & 11.35          & 19.82          & 32.75          & 14.08          & 24.94          & 41.80          & 17.76          & 19.77          & 32.80          & 14.14          \\
                                           & TimeMixer               & 16.21          & 26.63          & 12.11          & 19.12          & 31.17          & 14.84          & 22.34          & 36.94          & 17.67          & 18.78          & 30.81          & 14.48          \\
                                           & STGCN                   & 18.98          & 33.93          & 14.56          & 20.68          & 37.48          & 15.92          & 22.48          & 41.30          & 17.64          & 20.45          & 37.08          & 15.85          \\
                                           & MTGNN                   & 15.82          & 26.56          & 12.50          & 17.94          & 30.37          & 14.19          & 20.12          & 35.00          & 15.48          & 17.67          & 30.03          & 13.90          \\
                                           & CrossGNN                & 17.79          & 29.01          & 13.25          & 21.60          & 35.07          & 16.52          & 25.77          & 42.12          & 20.47          & 21.20          & 34.53          & 16.21          \\
                                           & Crossformer             & 15.33          & {\ul 25.60}    & 10.97          & {\ul 17.15}    & {\ul 28.59}    & {\ul 12.33}    & 19.61          & 32.79          & {\ul 14.95}    & {\ul 16.98}    & {\ul 28.34}    & {\ul 12.46}    \\
                                           & Card                    & 16.34          & 26.96          & 11.89          & 19.16          & 31.65          & 14.25          & 22.12          & 37.14          & 16.27          & 18.80          & 31.18          & 13.85          \\
                                           & RPMixer                 & 16.77          & 27.22          & 12.89          & 18.54          & 30.46          & 14.31          & 20.56          & 34.44          & 16.34          & 18.28          & 30.07          & 14.23          \\
                                           & BigST                   & 18.22          & 29.79          & 14.29          & 21.26          & 35.05          & 16.96          & 25.05          & 41.94          & 20.88          & 20.96          & 34.69          & 16.92          \\
                                           & Sumba                   & 15.38          & 25.67          & 11.08          & 17.24          & 28.89          & 12.58          & {\ul 19.54}    & {\ul 32.51}    & 15.11          & 17.02          & 28.42          & 12.60          \\
                                           & SOFTS                   & {\ul 15.19}    & 25.69          & {\ul 10.87}    & 17.60          & 29.77          & 13.01          & 20.99          & 35.66          & 16.07          & 17.55          & 29.62          & 13.08          \\
                                           & SBATransformer               & \textbf{14.07} & \textbf{24.11} & \textbf{9.74}  & \textbf{15.90} & \textbf{27.35} & \textbf{11.16} & \textbf{18.07} & \textbf{31.11} & \textbf{13.38} & \textbf{15.67} & \textbf{26.88} & \textbf{11.15} \\ \bottomrule
\end{tabular}}
\end{small}
\end{center}
\end{table}
\begin{table*}[]
\renewcommand{\arraystretch}{1.5}
\caption{Spatiotemporal series forecasting efficiency comparison. Time: training time (in seconds) per epoch. Mem: max memory used (in gigabytes) during training. - indicates out of memory.} 
\label{table:efficiency-analysis}
\begin{center}
\begin{small}
\resizebox{\linewidth}{!}{
\begin{tabular}{l|cc|cc|cc|cc|cc|cc|cc}
\toprule
Dataset      & \multicolumn{2}{c|}{SD} & \multicolumn{2}{c|}{GBA} & \multicolumn{2}{c|}{GLA} & \multicolumn{2}{c|}{CA} & \multicolumn{2}{c|}{WEST} & \multicolumn{2}{c|}{EAST} & \multicolumn{2}{c}{ALL} \\ \midrule
Metric       & Time        & Mem       & Time        & Mem        & Time       & Mem         & Time        & Mem       & Time         & Mem        & Time         & Mem        & Time        & Mem       \\ \midrule
PatchTST     & 204         & 30        & 348         & 94         & 309        & 154         & 383         & 172       & 692          & 66         & 743          & 186        & 529         & 190       \\
TimeMixer    & 191         & 40        & 340         & 132        & 311        & 188         & 551         & 180       & 645          & 91         & 756          & 170        & 587         & 157       \\
STGCN        & 615         & 39        & 1119        & 88         & 1168       & 138         & 1972        & 183       & 2275         & 88         & 3420         & 182        & 2828        & 181       \\
MTGNN        & 292         & 34        & 746         & 90         & 828        & 191         & 2749        & 182       & 1287         & 69         & 2966         & 185        & 2235        & 189       \\
CrossGNN     & 89          & 29        & 228         & 95         & 248        & 140         & 519         & 166       & 356          & 66         & 622          & 188        & 482         & 183       \\
iTransformer & 48          & 10        & 160         & 82         & 208        & 185         & -           & -         & 135          & 20         & 370          & 180        & 334         & 180       \\
Crossformer  & 49          & 8         & 107         & 24         & 125        & 39          & 432         & 91        & 153          & 15         & 245          & 58         & 369         & 70        \\
Card         & 225         & 36        & 407         & 117        & 367        & 192         & 707         & 170       & 767          & 83         & 909          & 159        & 663         & 148       \\
RPMixer      & 55          & 3         & 126         & 8          & 156        & 13          & 305         & 32        & 195          & 6          & 378          & 19         & 399         & 24        \\
Sumba        & 191         & 22        & 422         & 65         & 447        & 105       & 805         & 176       & 819          & 42         & 1359         & 158        & 906         & 178       \\
SOFTS        & 24          & 4         & 52          & 12         & 73         & 20          & 149         & 49        & 63           & 6          & 139          & 21         & 164         & 27        \\ \midrule
SBATransformer    & 55          & 5         & 130         & 19         & 167        & 29          & 283         & 68        & 73           & 6          & 211          & 35         & 198         & 35        \\ \bottomrule
\end{tabular}}
\end{small}
\end{center}
\end{table*}
Due to the space limitation of the main text, the full spatial-temporal series forecasting results are provided here. Table~\ref{table:full-forecasting-results-of-the-LargeST-datasets} contains detailed performance results for horizons 3, 6, 12,  and the average of all 12 horizons on the LargeST benchmark. Table~\ref{table:efficiency-analysis} shows detailed efficiency comparison results on the LargeST and PV-US benchmarks. Meanwhile, the ablation study results on the SD and GBA datasets is presented in Table~\ref{table:ablation-study-SD-GBA}. Figure~\ref{fig:effiency-analysis-CA} demonstrates intuitive efficiency comparison on the CA dataset.
\section{More Hyperparameters Sensitivity}
\label{sec:more-hyperparameters-sensitivity}
We assess the sensitivity of hyperparameters including the look-back time window \(T\) and the number of blocks \(L\).

\begin{figure}
    \centering
    \includegraphics[width=\textwidth]{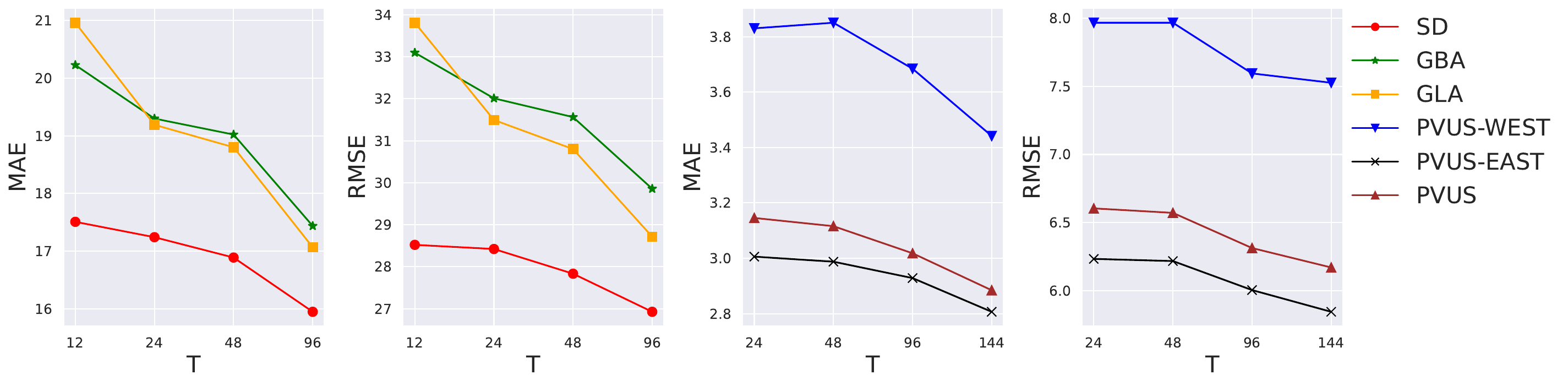}
    \vspace{-5mm}
    \caption{The sensitivity of our method to the look-back time window \(T\).}
    \label{fig:hyperparamter-sensitivity-T}
\end{figure}

\textbf{Look-back time window.}
Figure~\ref{fig:hyperparamter-sensitivity-T} demonstrates that increasing the look-back time window size  \(T\) enhances model performance for both datasets. However, as the input sequence length increases, the computational and memory costs of spatiotemporal GNNs surge rapidly. Thus, most existing models only utilize historical information within a short-term window for prediction, which significantly restricts their performance~\cite{han2024bigst}. In contrast, our model can not only achieve scalability in the spatial dimension but also accommodate a larger look-back time window to enhance performance.

\begin{figure}
    \centering
    \includegraphics[width=\textwidth]{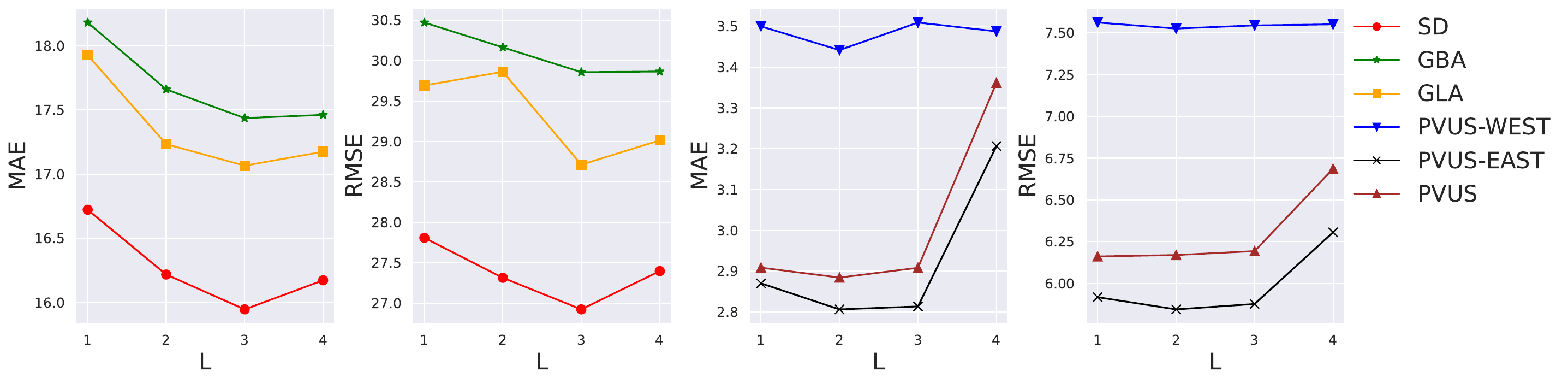}
    \vspace{-5mm}
    \caption{The sensitivity of our method to the number of blocks \(L\).}
    \label{fig:hyperparamter-sensitivity-L}
\end{figure}

\textbf{Number of blocks.}
As depicted in Figure~\ref{fig:hyperparamter-sensitivity-L}, model performance improves with the increase in the number of blocks \(L\) and achieves the best performance when \(L=3\) for LargeST datasets and \(L=2\) for PV-US datasets. Further increasing \(L\) does not lead to performance improvements. We propose that excessively deep layers can lead to overfitting, highlighting the necessity of a balanced architecture.
\section{Visualization}
\label{sec:visualization}
We visualize the local and global attention matrices of each block on the SD dataset to gain a deeper insight into our model and visualize the forecasting results on the GBA dataset to compare different models.

\begin{figure} \centering    
\subfigure[Local Attention Matrix] {    
\includegraphics[width=0.9\columnwidth]{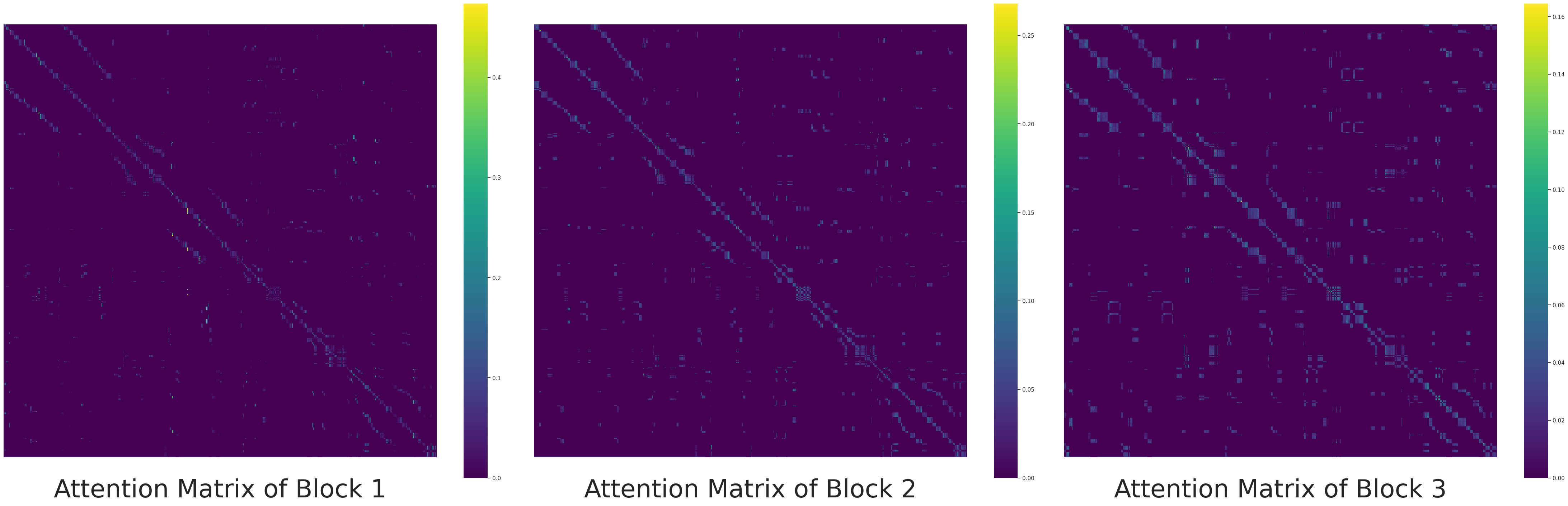}  
}     
\subfigure[Global Attention Matrix] {     
\includegraphics[width=0.9\columnwidth]{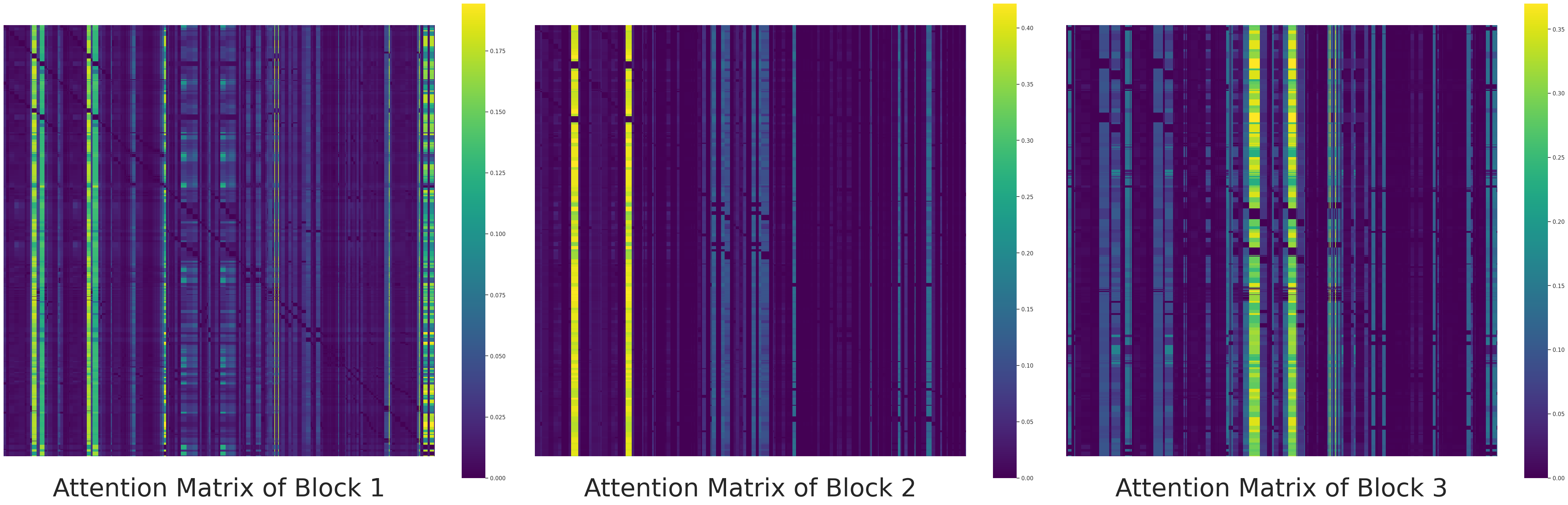}     
}    
\caption{Visualizations on the SD dataset.}     
\label{fig:attention-matrix-visualizations}   
\end{figure}

\textbf{Attention matrix.}
As shown in Figure~\ref{fig:attention-matrix-visualizations}-(a), the local attention matrix expands its receptive field as the block deepens and exhibits high-rank characteristics. In contrast, the global attention matrix exhibits low-rank characteristics, as shown in Figure~\ref{fig:attention-matrix-visualizations}-(b). We propose that this can be interpreted as the discovery of key hubs in a large-scale spatiotemporal network.

\begin{figure} 
\centering    
\subfigure[] {    
\includegraphics[width=0.9\columnwidth]{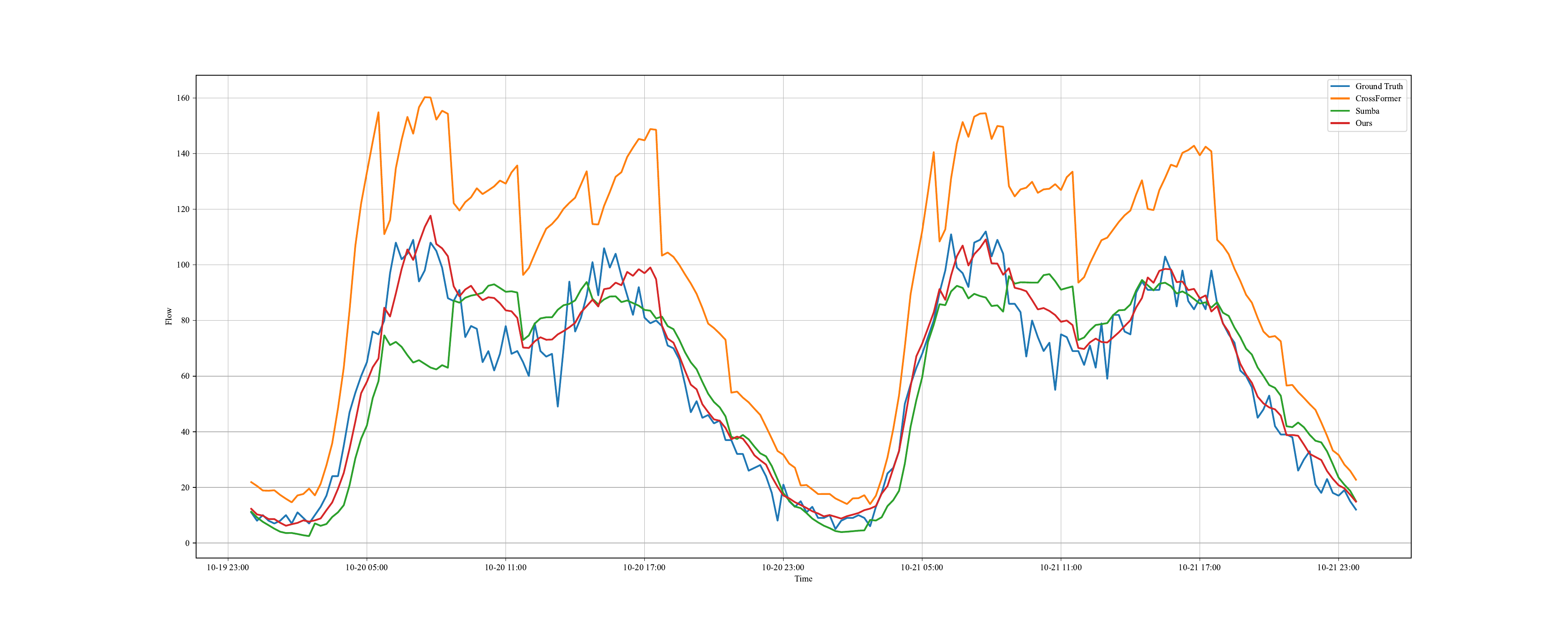}  
}     
\subfigure[] {     
\includegraphics[width=0.9\columnwidth]{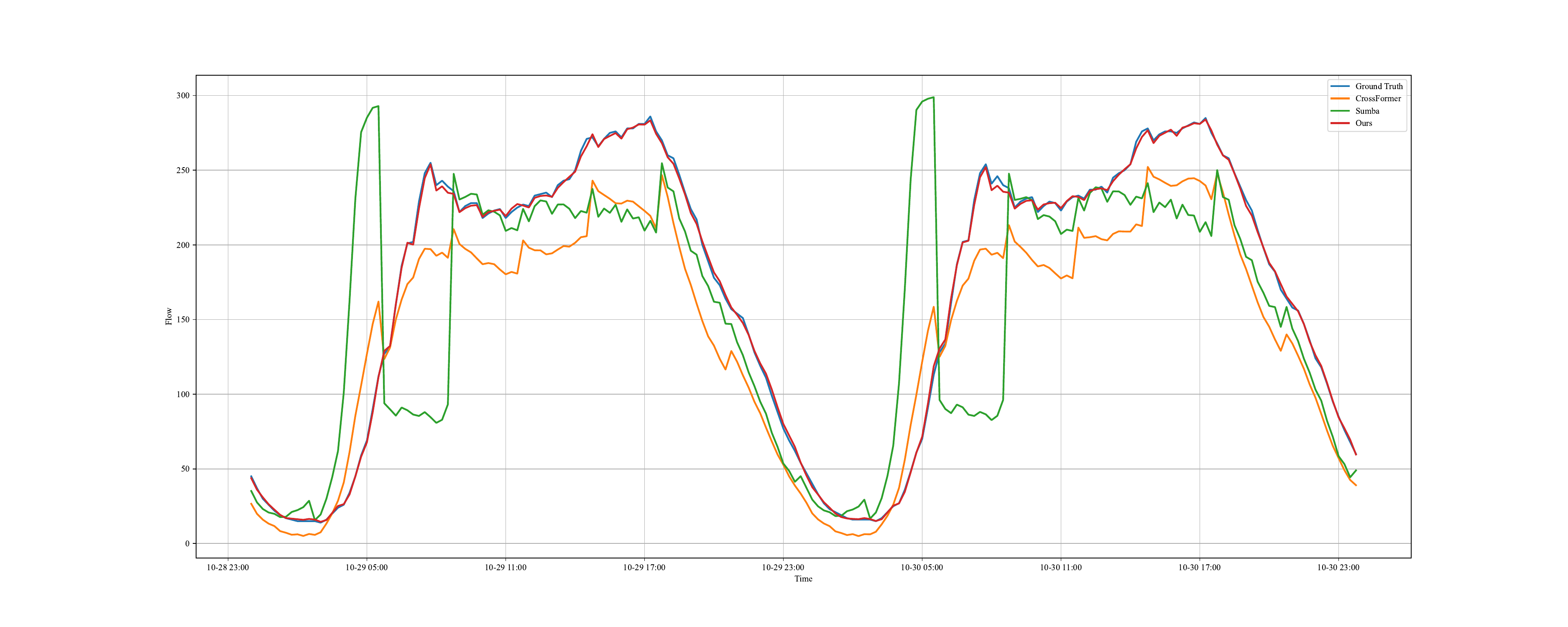}     
}    
\subfigure[] {    
\includegraphics[width=0.9\columnwidth]{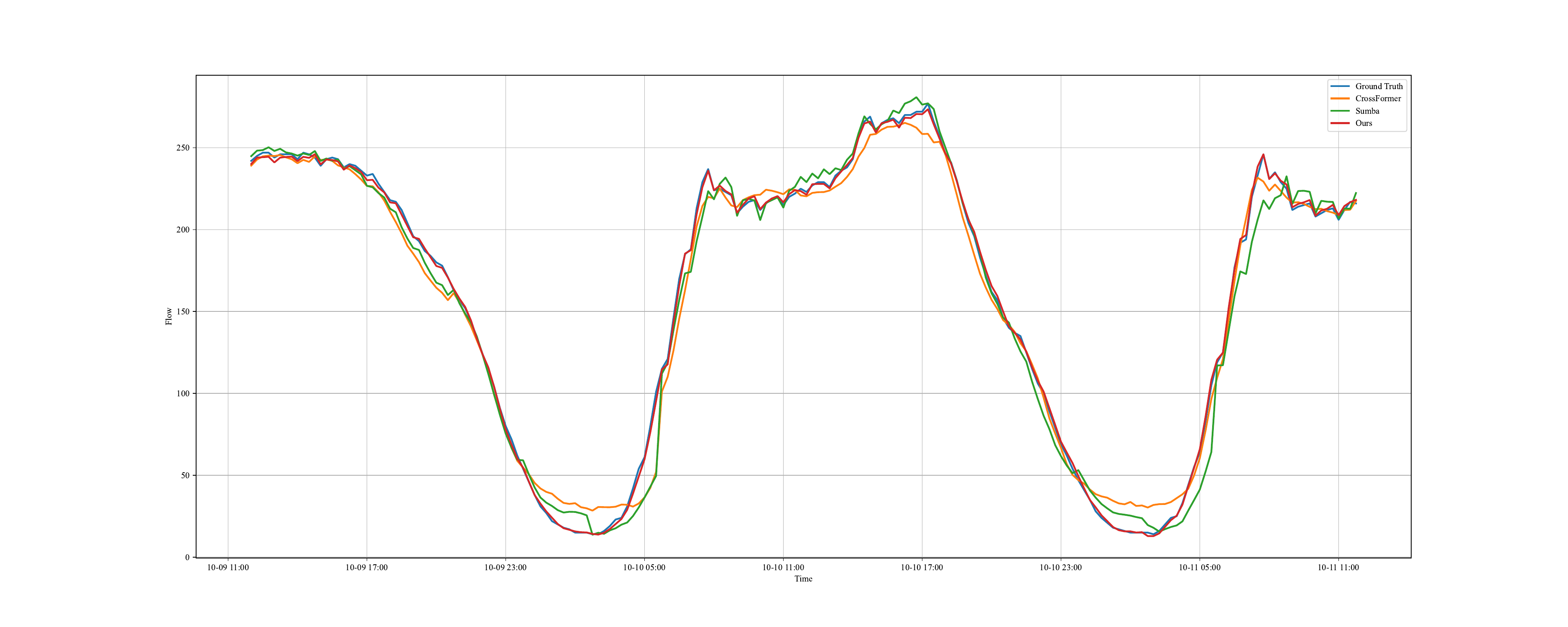}  
}  
\caption{Visualizations on the GBA dataset.}     
\label{fig:forecasting-results-visualizations}     
\end{figure}

\textbf{Forecasting result.}
As shown in Figure~\ref{fig:forecasting-results-visualizations}-(a), when there are small oscillations in traffic flows, our model generates smoother and more robust predictions compared to baselines. As shown in Figures~\ref{fig:forecasting-results-visualizations}-(b) and -(c), when the traffic flow remains stable, our model achieves better fitting results. This is because our model effectively captures local spatial correlation and can utilize the changes of neighboring nodes for more accurate forecasting.

These results demonstrate the robustness and effectiveness of our model in handling both commonly used normal-scale and extremely large real-world datasets.
\begin{table}[]
\caption{Ablation study of the SD and GBA datasets. The best results are highlighted in \textbf{bold}.} 
\label{table:ablation-study-SD-GBA}
\begin{center}
\begin{small}
\begin{tabular}{l|ccc|ccc}
\toprule
Dataset        & \multicolumn{3}{c|}{SD}                          & \multicolumn{3}{c}{GBA}                          \\ \midrule
Metric         & MAE            & RMSE           & MAPE(\%)       & MAE            & RMSE           & MAPE(\%)       \\ \midrule
w/o. PE        & 16.29          & 27.34          & 10.76          & 17.84          & 30.63          & 13.36          \\
w/o. Intra-Att & 17.47          & 28.91          & 11.83          & 18.71          & 31.11          & 15.15          \\
w/o. Inter-Att & 16.16          & 27.34          & 10.81          & 17.71          & 30.42          & 13.09          \\
w/o. MA        & 16.00          & 26.97          & 11.03          & 17.50          & 30.00          & 13.04          \\ \midrule
SBATransformer      & \textbf{15.95} & \textbf{26.93} & \textbf{10.75} & \textbf{17.44} & \textbf{29.86} & \textbf{13.03} \\ \bottomrule
\end{tabular}
\end{small}
\end{center}
\end{table}

\end{document}